\documentclass[10pt,twocolumn,letterpaper]{article}

\usepackage[pagenumbers]{cvpr} %

\usepackage{graphicx}
\usepackage{amsmath}
\usepackage{amssymb}
\usepackage{booktabs}
\usepackage{xcolor}
\usepackage{array,makecell}
\usepackage{caption}
\usepackage{graphicx}
\usepackage{wrapfig,lipsum}
\usepackage{enumitem}
\usepackage{amsmath}
\usepackage{bm}
\usepackage{arydshln}
\usepackage[accsupp]{axessibility}

\usepackage[pagebackref,breaklinks,colorlinks]{hyperref}

\usepackage[capitalize]{cleveref}
\crefname{section}{Sec.}{Secs.}
\Crefname{section}{Section}{Sections}
\Crefname{table}{Table}{Tables}
\crefname{table}{Tab.}{Tabs.}

\begin{document}

\title{A Closer Look at Rehearsal-Free Continual Learning}

\author{James Seale Smith\textsuperscript{1}\thanks{Correspondence to: James Seale Smith jamessealesmith@gatech.edu}\;, Junjiao Tian\textsuperscript{1}, Shaunak Halbe\textsuperscript{1}, Yen-Chang Hsu\textsuperscript{2}, Zsolt Kira\textsuperscript{1} \\
\normalsize
\textsuperscript{1}Georgia Institute of Technology, 
\textsuperscript{2}Samsung Research America 
}

\maketitle

\begin{abstract}
    Continual learning is a setting where machine learning models learn novel concepts from continuously shifting training data, while simultaneously avoiding degradation of knowledge on previously seen classes which may disappear from the training data for extended periods of time (a phenomenon known as the catastrophic forgetting problem). Current approaches for continual learning of a single expanding task (aka class-incremental continual learning) require extensive rehearsal of previously seen data to avoid this degradation of knowledge. Unfortunately, rehearsal comes at a cost to memory, and it may also violate data-privacy. Instead, we explore combining knowledge distillation and parameter regularization in new ways to achieve strong continual learning performance without rehearsal. Specifically, we take a deep dive into common continual learning techniques: prediction distillation, feature distillation, L2 parameter regularization, and EWC parameter regularization. We first disprove the common assumption that parameter regularization techniques fail for rehearsal-free continual learning of a single, expanding task. Next, we explore how to leverage knowledge from a pre-trained model in rehearsal-free continual learning and find that vanilla L2 parameter regularization outperforms EWC parameter regularization and feature distillation. Finally, we explore the recently popular ImageNet-R benchmark, and show that L2 parameter regularization implemented in self-attention blocks of a ViT transformer outperforms recent popular prompting for continual learning methods.
\end{abstract}

\section{Introduction}

Deep learning models for machine learning applications are typically trained offline on a large, static dataset. The model is then deployed to the real world with assumptions that the distribution of data it will encounter matches the distribution of data it was trained on. Unfortunately, this assumption does not hold for many applications because the model will encounter a natural distribution shift in the target data over time. These shifts lead to performance degradation, requiring that the model be replaced.

One way to replace a model is to collect additional training data, combine this new training data with the old training data, and then retrain the model from scratch. While this will guarantee high model performance, it is not practical for large-scale applications which may require long training times for the model. This may lead to \emph{high financial~\cite{justus2018predicting} and environmental~\cite{lacoste2019quantifying} costs} after numerous replacements. Instead, the preferred way is to update the model in the most efficient\footnote{W.r.t. to computation and/or memory, depending on the application.} manner possible. The simplest way to update the model is to train it on only the new training data. However, this leads to a phenomenon known as \textit{catastrophic forgetting}~~\cite{nguyen2019toward}, where the model overwrites previously acquired knowledge when learning the new data. This results in drastic performance degradation, or ``forgetting", over the previously learned training data distribution.

The study of catastrophic forgetting is referred to as \textbf{continual learning}. In this setting, a model sequentially learns from new task data while avoiding the catastrophic forgetting of previously seen data. 
This task data typically contain \textit{semantic} distribution shifts (e.g., we encounter new object classes) rather than \textit{covariate} distribution shifts 
\footnote{We note here that covariate distribution shifts have been studied in recent continual learning works~\cite{churamani2021domain,kundu2020class}, but this is not the focus of our paper. For more discussion on this comparison, the reader is referred to~\cite{van2019three}.}
(e.g., we encounter new lighting or background conditions). The goal of the continual learning problem is to find the most \textit{efficient} training strategy to update models which are sequentially trained on these task sequences. Strategies are typically evaluated on metrics such as task performance (e.g., classification accuracy for a classification problem),  computational efficiency (e.g., training time), and memory efficiency (e.g., number of parameters stored).

In this paper, we focus on continual learning over a single, expanding classification head. This is different from the multi-task continual learning setting, known as \textit{task-incremental} continual learning, where we learn separate classification heads for each task (and the task label is provided during inference)~\cite{hsu2018re}. Unfortunately, SOTA methods for continual learning without task labels require that a subset of the training data be stored or generated to mix in with future task data, a strategy referred to as \textbf{rehearsal}. Many applications are unable to store this data because they work with \emph{private user data that cannot be stored} long term. For example, some companies will collect user data to update the models in the short term (hours to day) but this data could have a timestamp and need deleting.

In this paper, we take a closer look at \emph{rehearsal-free} strategies for continual learning which do not store training data. Rather than propose a new method, we offer an interesting and impactful new perspective building on existing strategies. Specifically, we start by asking the question: \emph{what type of regularization (parameter-space or prediction-space) is best for rehearsal-free continual learning?} We provide analysis into \emph{how these methods forget} from a feature-drift perspective, and show that parameter regularization is most effective at reducing forgetting in the feature encoder while prediction distillation \emph{using multi-class sigmoid instead of softmax} is most effective for reducing forgetting and bias in the classifier head. 

Unfortunately, we show that the gap between rehearsal and rehearsal-free methods remains large. We conjecture that pre-training may help close this gap, leading us to our next question: \emph{what type of regularization (parameter-space or prediction-space) can best leverage a pre-trained model for rehearsal-free continual learning?} We surprisingly find that, while L2 regularization has low accuracy when the model is randomly initialized from scratch, it actually performs best in this pre-training setting and beats out more sophisticated methods, including recent prompting for continual learning methods~\cite{smith2022coda,wang2022learning,wang2022dualprompt}.

Finally, we show that a simple method derived from our findings can even outperform rehearsal-based methods on a standard continual learning benchmark. 
\emph{In summary, we make the following findings and contributions:}
\begin{enumerate}
[topsep=0pt,itemsep=-1ex,partopsep=1ex,parsep=1ex,leftmargin=*,labelindent=0pt]

    \item We provide a closer look into rehearsal-free continual learning with best practices, identifying that forgetting largely happens in the later layers. The most effective mitigation is through regularizing the final predictions when pre-training is not available.
    
    \item We extend the above investigations to the scenario where pre-training is available and find that regularizing parameters is more effective than regularizing predictions, pointing out the efficacy of methods can shift dramatically with continual learning problem settings.
    
    \item We achieve SOTA results in the rehearsal-free setting and even outperform recent SOTA prompting for continual learning methods~\cite{smith2022coda,wang2022learning,wang2022dualprompt}.

\end{enumerate}
\section{Background and Related Work}
\label{sec:rl}

\noindent
\textbf{Continual Learning}: 
Continual learning approaches can be organized into a few broad categories which are all useful depending on the problem setting and constraints. One group of approaches expand a model's architecture as new tasks are encountered; these are highly effective for applications where a model growing with tasks is practical~\cite{ebrahimi2020adversarial,lee2020neural,lomonaco2017core50,maltoni2019continuous,Rusu:2016}. We do not consider these methods because the model parameters grow with the number of tasks, but acknowledge that our contributions could be incorporated into these approaches. 

Another approach is to regularize the model with respect to past task knowledge while training the new task. This can either be done by regularizing the model in the  weight space (i.e., penalize changes to model parameters)~\cite{aljundi2017memory,ebrahimi2019uncertainty,kirkpatrick2017overcoming,titsias2019functional,zenke2017continual} or the prediction space (i.e., penalize changes to model predictions)~\cite{Ahn2021ssil,castro2018end,hou2018lifelong,lee2019overcoming,li2016learning}. Regularizing knowledge in the prediction space is done using \textit{knowledge distillation}~\cite{hinton2015distilling} and it has been found to perform better than model regularization based methods for continual learning when task labels are not given~\cite{lesort2019generative,van2018generative}.

Rehearsal with stored data~\cite{aljundi2019online, aljundi2019gradient,bang2021rainbow, chaudhry2018efficient,chaudhry2019episodic,Gepperth:2017,hayes2018memory,hou2019learning,Kemker:2017,Lopez-Paz:2017,Rebuffi:2016,robins1995catastrophic, rolnick2019experience,von2019continual} or samples from a generative model~\cite{kamra2017deep,kemker2018fear,ostapenko2019learning,Shin:2017,van2020brain} is highly effective when storing training data or training/saving a generative model is possible. Unfortunately for many machine learning applications, long-term storage of training data will violate data-privacy, as well as incurring a large memory cost. With respect to the generative model, this training process is much more computationally and memory intensive compared to a classification model and additionally may violate data legality concerns because using a generative model increases the chance of memorizing potentially sensitive data~\cite{nagarajan2018theoretical}. This motivates us to work on the important setting of \emph{rehearsal-free} approaches to mitigate catastrophic forgetting.

\noindent
\textbf{Online Rehearsal-Free Continual Learning}: 
Other works have looked at rehearsal-free continual learning from an online ``streaming" learning perspective using a frozen, pre-trained model~\cite{hayes:2019,lomonaco2020rehearsal}. While these works focus on efficient online learning from a fixed, frozen feature space, we instead analyze non-frozen models which are allowed to train "to convergence" on task data (as is common for offline continual learning~\cite{wu2019large}). Therefore, our setting is very different from these works.

\noindent
\textbf{Prototype-Based Approaches for Continual Learning}:
Prototypes can be leveraged for continual learning as a means to avoid catastrophic forgetting without storing data. Recent methods learn a feature space for prototypes with approaches such as learning an embedding network~\cite{yu2020semantic} or leveraging strong augmentations for self-supervised learning~\cite{wu2021striking,zhu2021prototype}. While learning prototypes in an embedding network~\cite{yu2020semantic} can better mitigate forgetting compared to cross-entropy classification, we avoid such approaches because training an embedding network with metric learning can often be a hard challenge~\cite{zhu2021prototype}. While leveraging strong self-supervision to augment data and prototypes can achieve SOTA performance for rehearsal-free continual learning~\cite{wu2021striking,zhu2021prototype}, it is not clear if the performance increase is due to mitigating forgetting versus having generally better features due to an expanded dataset of strong data augmentations~\cite{cubuk2020randaugment}. Additionally, these approaches \emph{require} a large first-task to learn a strong initial feature space (which is not always valid). In summary, while these advanced strategies perform well in the absence of stored data, we instead offer our work as a different perspective on simple, existing, widely-adopted strategies rather than a complex, SOTA method which requires additional assumptions (e.g., having a large first task).

\noindent
\textbf{Rehearsal-Free Continual Learning}: 
Recent works learn prompts within a frozen, pre-trained transformer model for continual learning~\cite{wang2022learning,wang2022dualprompt,smith2022coda}. While effective, this approach assumes that the data within the continual learning sequence can be separated with a pre-trained encoder; because this assumption is often invalid, it is still strongly desired to understand how fine-tuning based approaches forget in the rehearsal-free setting. Other works propose producing images for rehearsal using deep-model inversion~\cite{choi2021dual,smith2021abd,yin2020dreaming}. While these methods perform well compared to generative modeling approaches and simply rehearsal from a small number of stored images, we argue that these methods have similar risks to generative approaches. Specifically, model-inversion is a slow process associated with high computational costs in the continual learning setting~\cite{smith2021abd} and inverting images from a trained model can also violate the same data-privacy concerns~\cite{kaissis2021end}. This motivates us to ask: ``how can we \emph{entirely eliminate rehearsal including stored, trained, or inverted training data?}" %

\section{Preliminaries}

\noindent
\textbf{Continual Learning}: In continual learning, a model is shown labeled data corresponding to $M$ semantic object classes $c_1, c_2, \dots, c_M$ over a series of $N$ tasks corresponding to non-overlapping subsets of classes. We use the notation $\mathcal{T}_n$ to denote the set of classes introduced in task $n$, with $|\mathcal{T}_n|$ denoting the number of object classes in task $n$. Each class appears in only a single task, and the goal is to incrementally learn to classify new object classes as they are introduced while retaining performance on previously learned classes. To describe our inference model, we denote $\theta_{i,n}$ as the model $\theta$ at time $i$ that has been trained with the classes from task $n$. For example, $\theta_{n,1:n}$ refers to the model trained during task $n$ and the linear classification heads associated with all tasks up to and including class $n$. We drop the second index when describing the model trained during task $n$ with all linear classification heads (for example, $\theta_{n}$).

In this paper, we deal with the \textit{class-incremental continual learning setting} rather than the \textit{task-incremental continual learning} setting. Class-incremental continual learning is challenging because the learner must support classification across all classes seen up to task $n$~\cite{hsu2018re} (i.e., \textit{no task labels are provided to the learner during inference}). Task-incremental continual learning is a simpler \textit{multi-task} setting where the task labels are given during both training and inference. 
\begin{figure*}[t]
    \centering
    \includegraphics[width=0.9\textwidth]{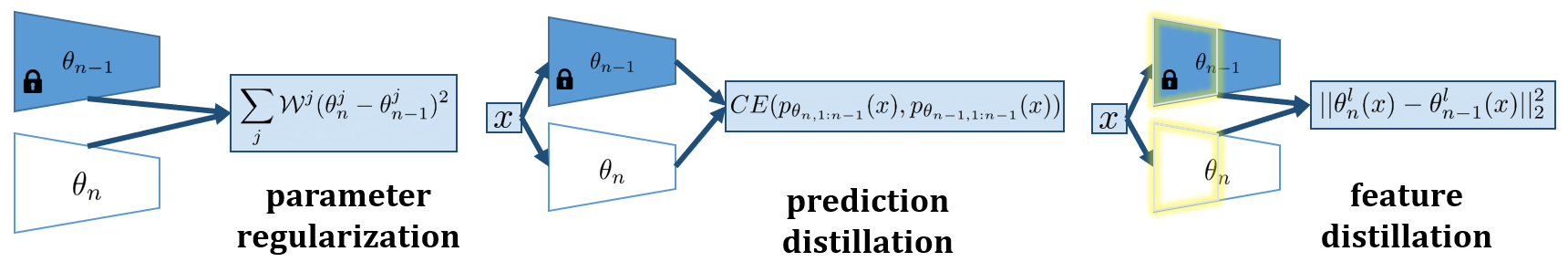}
    \caption{
    \textbf{Three ways to transfer knowledge from a checkpoint model in continual learning}. Parameter regularization penalizes changes in the \emph{model parameter space}. This can be weighted by the Fisher Information Matrix (e.g., EWC~\cite{kirkpatrick2017overcoming}) or with no weighting (e.g., L2 regularization). Prediction distillation (e.g., LwF~\cite{li2016learning}) penalizes features in the \emph{model prediction space} with respect to some data $x$, whereas feature distillation penalizes features in the \emph{model intermediate feature space} with respect to some data $x$. In the case where $x$ is strictly from the new task, both distillation methods are ``rehearsal-free".
    }
    \vspace{-.2cm}
    \label{fig:losses}
\end{figure*}

\section{Rehearsal-Free Regularization}

When training on a new task $n$, the key to mitigating forgetting is to transfer knowledge from a ``checkpoint" model, $\theta_{n-1}$ (which is copied and frozen at the end of task $n-1$), into the model being updated, $\theta_n$. In this section, we first review three classic ways to transfer knowledge in continual learning which can be described as ``rehearsal-free". These approaches are visualized in Figure~\ref{fig:losses}, and we encourage the reader to refer back to Figure~\ref{fig:losses} throughout reading this section. We then argue that one of these methods, prediction distillation, is more important for transferring knowledge from a model's \emph{classifier}, whereas the other two methods, parameter regularization and feature distillation, are more important for transferring knowledge from  a model's \emph{feature encoder}. We will use this section as a foundation to motivate and understand the findings presented in Section~\ref{sec:exp}.

\subsection{Parameter Space Regularization}

One of the earliest approaches for continual learning, EWC, proposed regularizing the model in the \emph{model parameter space}~\cite{kirkpatrick2017overcoming}. At a high level, this approach searches for a solution in each new task that lies within the weight space of solutions to the previous tasks. This is done by calculating the L2 distance between each model parameter in $\theta_{n-1}$ and each model parameter in $\theta_n$, or:
\begin{equation}
    \mathcal{L}_{ewc} =  \sum_{j=1}^{N_{params}} F_{n-1}^{jj} \left( \theta_n^{j} - \theta_{n-1}^{j} \right)^2 
    \label{eq:ewc}
\end{equation}
where $F_{n-1}^{jj}$ is the $j^{th}$ diagonal element of the ${n-1}^{th}$ Fisher information matrix $F_{n-1}$, which is calculated using the data and loss function in task $n-1$. We refer to this approach as \textbf{EWC} throughout this paper. Observe that if $F$ is given as the identify matrix, $\mathcal{L}_{ewc}$ simply becomes L2 regularization between the model parameters. We will analyze this approach as well and refer to it as simply \textbf{L2} for the rest of this paper. A strong advantage of L2 regularization versus EWC regularization is that L2 regularization can be applied in the absence of an importance-weighting matrix (e.g., L2 can be applied in the first task of a continual learning sequence in the presence of pre-training to retain the pre-trained knowledge).  While the original work shows that using the identify matrix for $F$ hurts performance, we will show later that L2 can actually outperform EWC under certain continual learning settings.

\subsection{Feature Space Regularization}

Another approach for continual learning is to leverage \emph{knowledge distillation} from $\theta_{n-1}$ to regularize the learning of $\theta_n$. This was first introduced for continual learning in the Learning without Forgetting (LwF)~\cite{li2016learning} method as knowledge distillation \emph{in the prediction space}. We will refer to this as \textbf{PredKD} throughout the paper.

Let us denote $p_{\theta}(y \mid x)$ as the predicted class distribution produced by model $\theta$ for input $x$. Using this notation, the loss function for PredKD is defined as:
\begin{equation}
    \mathcal{L}_{PredKD} = CE(p_{\theta_{n,1:n-1}}(x), p_{\theta_{n-1,1:n-1}}(x))
\label{eq:pred-kd}
\end{equation}
where $CE$ is the standard cross-entropy loss. Knowledge can also be distilled in the \emph{feature space} instead of the \emph{prediction space}. The intuition here is to directly align the models' feature space so that the feature space does not drift far from the previous checkpoint solution.  We will refer to this as \textbf{FeatureKD} throughout the paper with a loss function given as:
\begin{equation}
    \mathcal{L}_{FeatKD} = ||\theta_{n}^{l}(x) - \theta_{n-1}^{l}(x) ||^2_2
\label{eq:feature-kd}
\end{equation}
Notice that, since we do not generate class predictions $p_{\theta}(y \mid x)$ at the intermediate feature space, we instead minimize the squared error.

\subsection{Task-Bias}

Another continual learning phenomena that exists in the absence of task labels during inference is \emph{task bias} towards recent task data. This is typically mitigated with solutions relying on rehearsal data~\cite{wu2019large,Ahn2021ssil}. Since we cannot reduce task bias with rehearsal data in our setting, we borrow from the rehearsal-free continual learning method LWF.MC~\cite{Rebuffi:2016} and use \emph{sigmoid binary cross-entropy classification loss} (referred to as BCE) instead of the typical \emph{softmax cross-entropy classification loss} (we note here that this is nearly equivalent to using the "labels trick" from \cite{zeno2018task}). The intuition here is that \emph{softmax classification without rehearsal data results in a strong bias against the previously seen classes} because minimizing this loss reduces the magnitude of the old classes' corresponding logit outputs. We will show that the BCE classifier boosts the methods EWC, L2, and FeatKD into competitive SOTA approaches, despite having been previously reported to ``fail" in the continual learning setting when task labels are not present~\cite{hsu2018re}.
\section{Experiments}
\label{sec:exp}

\begin{table*}[t]
\caption{\textbf{Ablation results (\%) on 10 task CIFAR-100.} $A_{1:N}$ gives the final task accuracy, $F_N^G$ gives the average \emph{global} forgetting, and $F_L^G$ gives the average \emph{local} forgetting. \emph{BCE} refers to binary cross-entropy loss whereas \emph{Soft} refers to softmax cross-entropy loss. We report the mean over 3 trials.}
\vspace{-.3cm}
\begin{subtable}[h]{0.49\textwidth}
\centering
\caption{parameter regularization}
\resizebox{1.0\linewidth}{!}{
    \begin{tabular}{c|c|c|c} 
    \hline
    \rule{0pt}{10pt} Method & $A_{1:N}$ ($\uparrow$) & $F_N^G$ ($\downarrow$) & $F_N^L$ ($\downarrow$) \\
    \hline
    
    Upper-Bound  & $ 56.2 $ & $ 0.0 $ & $ 0.0 $  \\ 
    \hline

    PredKD+EWC (BCE)  & $ \bm{22.7} $ & $ \bm{-0.7} $ & $ \bm{7.8} $  \\ 
    EWC (BCE)  & $ 7.7 $ & $ 64.0 $ & $ 58.5 $  \\ 
    PredKD+EWC (Soft)  & $ 7.3 $ & $ 44.4 $ & $ 56.8 $  \\ 
    EWC (Soft)  & $ 7.3 $ & $ 52.7 $ & $ 57.8 $  \\

    \hline
    \end{tabular}
    }

\end{subtable}
\hfill
\begin{subtable}[h]{0.49\textwidth}
\centering
\caption{feature regularization}
\resizebox{1.0\linewidth}{!}{
    \begin{tabular}{c|c|c|c} 
    \hline
    \rule{0pt}{10pt} Method & $A_{1:N}$ ($\uparrow$) & $F_N^G$ ($\downarrow$) & $F_N^L$ ($\downarrow$) \\
    \hline
    
    Upper-Bound  & $ 56.2 $ & $ 0.0 $ & $ 0.0 $  \\ 
    \hline

    PredK +FeatKD (BCE)  & $ \bm{19.1} $ & $ \bm{7.0} $ & $ \bm{26.5} $  \\ 
    FeatKD (BCE)  & $ 8.2 $ & $ 66.5 $ & $ 58.2 $  \\ 
    PredKD+FeatKD (Soft)  & $ 8.2 $ & $ 54.2 $ & $ 60.5 $  \\ 
    FeatKD (Soft)  & $ 8.5 $ & $ 63.9 $ & $ 61.8 $  \\

    \hline
    \end{tabular}
    }
\label{tab:resnet18_tentask_ablate}
\end{subtable}

\end{table*}

In this section, we take a closer look at EWC, L2, PredKD, and FeatKD in the rehearsal-free continual learning setting. We analyze performance of these four losses in addition to both i) a naive model trained with classification loss only (referred to as \emph{naive}) and ii) an upper bound model trained with the joint training data from all tasks (referred to as \emph{upper-bound}). 
We first provide benchmark results on the CIFAR-100 dataset~\cite{krizhevsky2009learning} which contains 100 classes of 32x32x3 images. We train with a 18-layer ResNet~\cite{he2016deep} for 250 epochs using Adam optimization; the learning rate is set to 1e-3 and is reduced by 10 after 100, 150, and 200 epochs. We use a weight decay of 0.0002 and batch size of 128. Importantly, we do not tune our hyperparameters (i.e., the loss weights) on the full task set because tuning hyperparameters with hold out data from all tasks may violate the principle of continual learning that states each task is visited only once~\cite{van2019three}. Instead, we tuned our hyperparameters (including the loss weight for each approach) (using a half-decade linear sweep from $1e-3$ to $1e2$) on a small task sequence of each dataset.

\noindent
\textbf{Evaluation Metrics:} We evaluate methods using 
final accuracy, or the accuracy with respect to all past classes after having seen all $N$ tasks (referred to as $A_{N,1:N}$). Specifically, we have:
\begin{equation}
    A_{i,n} = \frac{1}{|\mathcal{D}_n^{test}|} \sum_{(x,y)\in\mathcal{D}_n^{test}} \bm{1}(\hat{y}(x,\theta_{i,n}) = y \mid \hat{y} \in \mathcal{T}_n)
\end{equation}
Note that $A_{i,n}$ gives the local task accuracy (i.e., inference in \emph{task-incremental} learning where the task label is given, used to calculate local forgetting $F_N^L$ below) and $A_{i,1:n}$ gives the global task accuracy (i.e., the accuracy when the task label is unknown, used to calculate global forgetting $F_N^G$ below). For the final task accuracy in our results, we will denote $A_{N,1:N}$ as simply $A_{1:N}$. We also measure: 
(I) global forgetting, or the measurement of average decrease in performance on task $n$ with respect to the \textit{global} task where no task label is given (referred to as $F_N^G$); and 
(II) local forgetting, or the measurement of average decrease in performance on task $n$ with respect to the \textit{local} task where the task index is given (referred to as $F_N^L$). Global forgetting is taken from \cite{lee2019overcoming} and given as:
\begin{equation} 
    F_N^G = \frac{1}{N-1} \sum_{i=2}^N \sum_{n=1}^{i-1} \frac{|\mathcal{T}_{n}|}{|\mathcal{T}_{1:i}|} (R_{n,n} - R_{i,n})
\end{equation}
where:
\begin{equation}
    R_{i,n} = \frac{1}{|\mathcal{D}_n^{test}|} \sum_{(x,y)\in\mathcal{D}_n^{test}} \bm{1}(\hat{y}(x,\theta_{i,1:n}) = y)
\end{equation}
and local forgetting is taken from \cite{Lopez-Paz:2017} and given as:
\begin{equation} 
    F_N^L = \frac{1}{N-1}  \sum_{n=1}^{N-1}  (A_{N,n} - A_{n,n})
\end{equation}

\subsection{Rehearsal-Free Continual Learning}
\label{sec-5a}

\begin{table}[t]
\caption{\textbf{Results (\%) on 10 task CIFAR-100} using BCE classification. $A_{1:N}$ gives the final task accuracy, $F_N^G$ gives the average \emph{global} forgetting, and $F_L^G$ gives the average \emph{local} forgetting. We report the mean over 3 trials.}
\label{tab:resnet18_tentask:a}
\centering
\begin{tabular}{c|c|c|c} 
\hline
\rule{0pt}{10pt} Method & $A_{1:N}$ ($\uparrow$) & $F_N^G$ ($\downarrow$) & $F_N^L$ ($\downarrow$) \\
\hline

Upper-Bound  & $ 56.2 $ & $ 0.0 $ & $ 0.0 $  \\ 
\hline

Naive  & $ 8.6 $ & $ 71.0 $ & $ 63.4 $  \\ 
PredKD  & $ \bm{25.2} $ & $ 3.2 $ & $ 27.2 $  \\ 
PredKD + FeatKD  & $ 19.1 $ & $ 7.0 $ & $ 26.5 $  \\ 
PredKD + EWC  & $ 22.7 $ & $ \bm{-0.7} $ & $ \bm{7.8} $  \\ 
PredKD + L2  & $ 21.6 $ & $ 1.6 $ & $ 15.4 $  \\

\hline
\end{tabular}

\end{table}

\begin{figure*}[!ht]
    \centering
    \begin{subfigure}{0.44\textwidth}
        \centering
        \includegraphics[width=\textwidth,trim={0 0 0 1.3cm},clip]{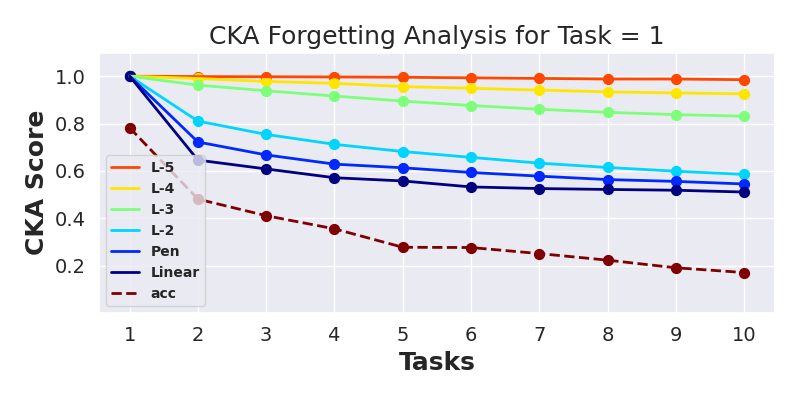}
        \vspace{-0.7cm}
        \caption{PredKD}
    \end{subfigure}
    \hspace{0.5cm}
    \begin{subfigure}{0.44\textwidth}
        \centering
        \includegraphics[width=\textwidth,trim={0 0 0 1.3cm},clip]{figures/plots/resnet18_10-task_lwf-mc_task-1.png}
        \vspace{-0.7cm}
        \caption{PredKD + FeatKD}
    \end{subfigure}
    
    \vspace{0.2cm}
    
    \begin{subfigure}{0.44\textwidth}
        \centering
        \includegraphics[width=\textwidth,trim={0 0 0 1.3cm},clip]{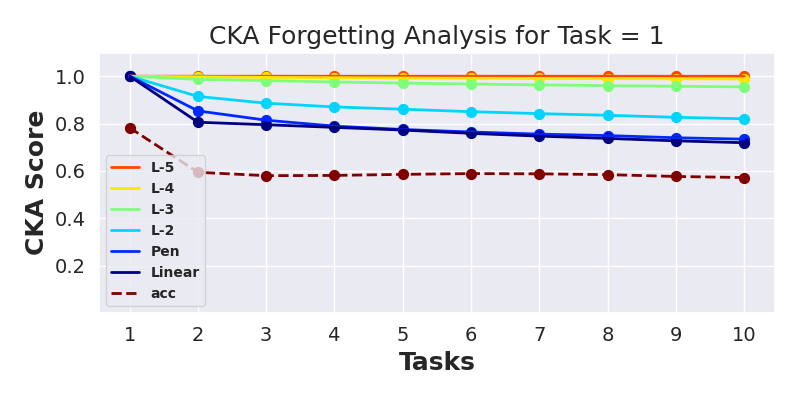}
        \vspace{-0.7cm}
        \caption{PredKD + EWC}
    \end{subfigure}
    \hspace{0.5cm}
    \begin{subfigure}{0.44\textwidth}
        \centering
        \includegraphics[width=\textwidth,trim={0 0 0 1.3cm},clip]{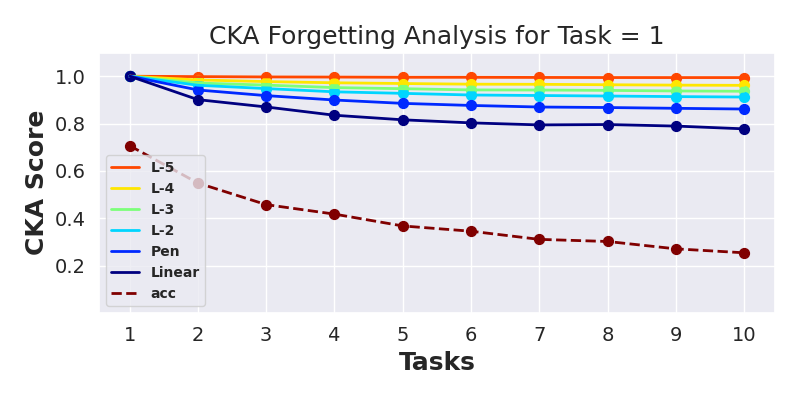}
        \vspace{-0.7cm}
        \caption{PredKD + L2}
    \end{subfigure}
    \vspace{-.15cm}
    \caption{
    \textbf{CKA Analysis on task-1 forgetting} for continual learning on CIFAR-100 for 10 tasks. \emph{Linear} refers to the output of the linear layer, \emph{pen} refers to the output of the penultimate layer, and \emph{L-2..4} refers to the outputs at second-to-last, third-to-last, etc. layers. \emph{Acc} refers to $A_{n,1:n}$, where $n$ is the task number (i.e., x axis).
    }
    \vspace{-.1cm}
    \label{fig:resnet18_tentask_cka}
\end{figure*}

We start by analyzing performance on a 10 task sequence from CIFAR-100. Here, our model is shown 10 different tasks derived of 10 classes per task from the CIFAR-100 dataset. We use loss weights of $\{1e1,5e-1,1,5\}$ for EWC, L2, PredKD, and FeatKD. Our first finding is that \textbf{PredKD and BCE are foundational for rehearsal-free continual learning}. In Table~\ref{tab:resnet18_tentask_ablate}, we tease apart two approaches which mitigate ``feature drift": EWC and FeatKD\footnote{Here, we leave out L2 given that it is a special case of EWC.}. For the two sides of this table, the top rows refer to the feature-drift method (EWC or FeatKD) using BCE when combined with PredKD. Below, we ablate the two methods separately, showing performance when i) PredKD is removed, ii) BCE is replaced with softmax classification, and iii) when PredKD is removed \emph{and} BCE is replaced with softmax classification.

The bottom row demonstrates that vanilla EWC and FeatKD fail for continual learning (poor $A_{1:N}$) yet do reasonably well in mitigating local forgetting $F_N^L$ when a softmax PredKD is added (i.e., they perform well for task-incremental learning where the task label is given)~\cite{hsu2018re}. The deeper finding here is that both EWC and FeatKD perform well at regularizing the \emph{feature drift} yet fail at regularizing/debiasing the \emph{classifier head}. As motivated in the prior section, we see that a significant jump in performance is achieved when combining \emph{both BCE and PredKD}.

In order to closer examine the effects of parameter regularization and feature distillation on catastrophic forgetting, we consider the following approaches i) PredKD\footnote{Notice that this is equivalent to the LwF.MC method from~\cite{Rebuffi:2016}.}, ii) PredKD + FeatKD, iii) PredKD + EWC, and iv) PredKD + L2. Specifically, we want to understand \emph{where forgetting is occurring} in these methods. We borrow the practice from ~\cite{ramasesh2020anatomy} and look at the centered kernel alignment (CKA) similarity between feature representations over time for different layers in the model (higher is better). In Figure~\ref{fig:resnet18_tentask_cka}, we see the CKA similarity score plotted for each layer in each model across tasks for the task-1 data. The x-axis at task $n$ gives the CKA similarity score between features evaluated on task-1 holdout data from $\theta_1$ versus the features generated on task-1 holdout data from $\theta_n$. We calculate the CKA at the following layers: \emph{Linear}, or the output of the linear layer; \emph{pen}, or the output of the penultimate layer; and \emph{L-2,L-3,L-4}, or the outputs at the second-to-last, third-to-last, and fourth-to-last layers. \emph{Acc} refers to the accuracy $A_{n,1:n}$, where $n$ is the task number (i.e., x axis). 
We can interpret these scores with the full results in Table~\ref{tab:resnet18_tentask:a}. For this experiment, we see that PredKD converges to the highest final accuracy, while PredKD + EWC has the lowest forgetting. Why is this? When we look at Figure~\ref{fig:resnet18_tentask_cka}, we see an \emph{trade-off} between retaining task 1 similarity across all layers versus final accuracy. Specifically, adding the parameter regularization losses (EWC and L2) induce \emph{low forgetting} but at the cost of \emph{low plasticity} (i.e., the ability to learn a new task). One surprising finding (which extends to the rest of this paper) is that FeatKD reduces the final accuracy without gaining any improvements in forgetting. \emph{In summary, the main takeaways from this section are that:} \textbf{1) PredKD and BCE create a strong baseline for rehearsal-free continual learning} and\textbf{ 2) Parameter regularization in addition to this baseline reduces forgetting, but at the expense of low plasticity and therefore low final accuracy}.
\subsection{How to Leverage Pre-Trained Models}
\begin{table}[t]
\caption{\textbf{Results (\%) on 10 task CIFAR-100 \emph{leveraging pre-training}} and BCE classification. $A_{1:N}$ gives the final task accuracy, $F_N^G$ gives the average \emph{global} forgetting, and $F_L^G$ gives the average \emph{local} forgetting. We report the mean over 3 trials.}
\vspace{-3mm}
\centering
\label{tab:resnet18_tentask:b}
\begin{tabular}{c|c|c|c} 
\hline
\rule{0pt}{10pt} Method & $A_{1:N}$ ($\uparrow$) & $F_N^G$ ($\downarrow$) & $F_N^L$ ($\downarrow$) \\
\hline

Upper-Bound  & $ 56.2 $ & $ 0.0 $ & $ 0.0 $  \\ 
\hline

Naive  & $ 8.5 $ & $ 75.6 $ & $ 67.6 $  \\ 
PredKD  & $ 26.6 $ & $ 3.9 $ & $ 34.3 $  \\ 
PredKD + FeatKD  & $ 23.5 $ & $ 7.5 $ & $ 25.3 $  \\ 
PredKD + EWC  & $ 31.1 $ & $ \bm{-0.4} $ & $ \bm{12.2} $  \\ 
PredKD + L2  & $ \bm{35.6} $ & $ 1.0 $ & $ 15.0 $  \\

\hline
\end{tabular}

\end{table}

\begin{figure*}[t]
    \centering
    \begin{subfigure}{0.44\textwidth}
        \centering
        \includegraphics[width=\textwidth,trim={0 0 0 1.3cm},clip]{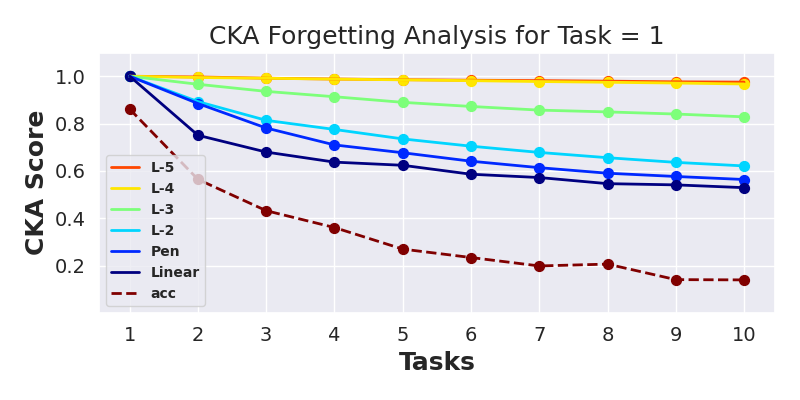}
        \vspace{-0.7cm}
        \caption{PredKD}
    \end{subfigure}
    \hspace{0.5cm}
    \begin{subfigure}{0.44\textwidth}
        \centering
        \includegraphics[width=\textwidth,trim={0 0 0 1.3cm},clip]{figures/plots/resnet18-pt_10-task_lwf-mc_task-1.png}
        \vspace{-0.7cm}
        \caption{PredKD + FeatKD}
    \end{subfigure}
    
    \vspace{0.2cm}
    
    \begin{subfigure}{0.44\textwidth}
        \centering
        \includegraphics[width=\textwidth,trim={0 0 0 1.3cm},clip]{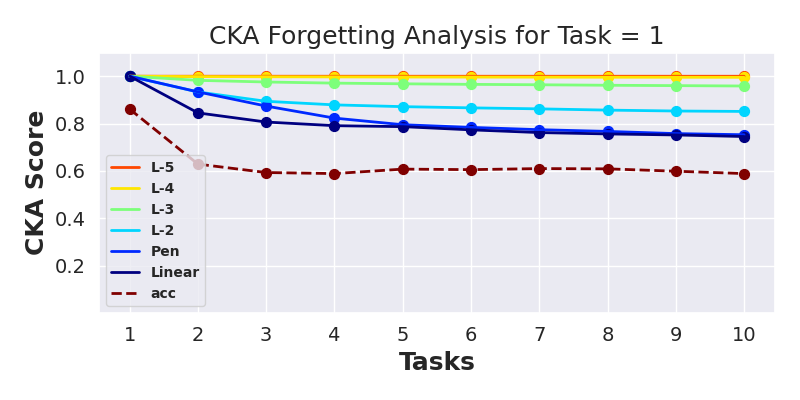}
        \vspace{-0.7cm}
        \caption{PredKD + EWC}
    \end{subfigure}
    \hspace{0.5cm}
    \begin{subfigure}{0.44\textwidth}
        \centering
        \includegraphics[width=\textwidth,trim={0 0 0 1.3cm},clip]{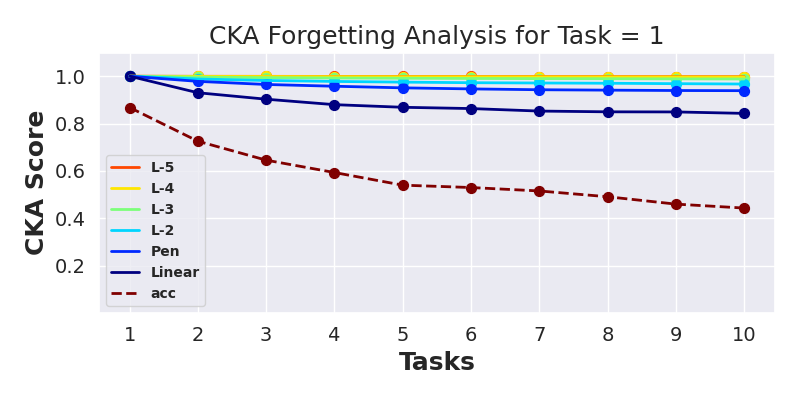}
        \vspace{-0.7cm}
        \caption{PredKD + L2}
    \end{subfigure}
    \caption{
    \textbf{CKA Analysis on task-1 forgetting} for continual learning on CIFAR-100 for 10 tasks with \textbf{model pretraining} on ImageNet1k. \emph{Linear} refers to the output of the linear layer, \emph{pen} refers to the output of the penultimate layer, and \emph{L-2..4} refers to the outputs at second-to-last, third-to-last, etc. layers. \emph{Acc} refers to $A_{n,1:n}$, where $n$ is the task number (i.e., x axis).
    }
    \vspace{-.2cm}
    \label{fig:resnet18-pt_tentask_cka}
\end{figure*}
\label{sec-5b}
\begin{figure*}[t]
    \centering
    \begin{subfigure}{0.42\textwidth}
        \centering
        \includegraphics[width=\textwidth,trim={0 0 0 0},clip]{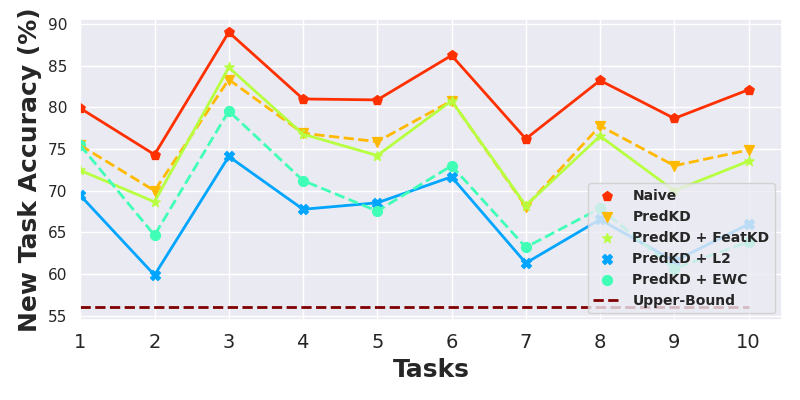}

        \caption{no pre-training}
    \end{subfigure}
    \hspace{0.5cm}
    \begin{subfigure}{0.42\textwidth}
        \centering
        \includegraphics[width=\textwidth,trim={0 0 0 0},clip]{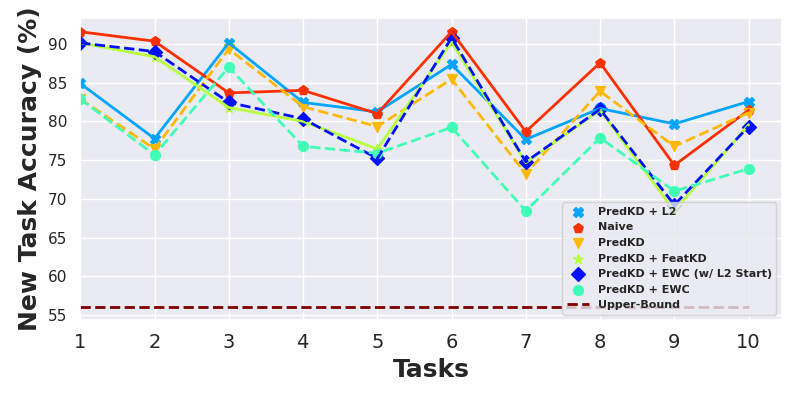}

        \caption{ImageNet1k pre-training}
    \end{subfigure}
    \caption{
    \textbf{Most recent task accuracy versus tasks seen}. Here, the accuracy at task $n$ is the \emph{local task accuracy} $A_{n,n}$ (as opposed to $A_{n,1:n}$), which we use to represent the ability to learn each new tasks. These plots demonstrate that the methods with ImageNet1k pre-training can achieve higher performance on new tasks (i.e. they require less plasticity to adapt to new tasks) compared to the methods with no pre-training.
    }
    \vspace{-0.25cm}
    \label{fig:resnet18_tentask_plasctiicy}
\end{figure*}
\looseness=-1
Because the gap between SOTA and the upper bound for rehearsal-free continual learning remains large, we explore leveraging pre-trained models. Specifically, we ask the question: \emph{what type of regularization (parameter-space or prediction-space) can best leverage model pre-training (from an auxiliary dataset) for rehearsal-free continual learning?} We note here that our work differs from~\cite{ramasesh2021effect} in that we \emph{analyze the effect of regularization on forgetting in pre-trained models} rather than \emph{show that pre-trained models are more robust to forgetting}. We repeat our experiments from the previous section, but this time our model is initialized with ImageNet~\cite{russakovsky2015imagenet} pre-training. We use loss weights of $\{1e2,1e-1,5,5\}$  for EWC, L2, PredKD, and FeatKD. The main results are found in Table~\ref{tab:resnet18_tentask:b}. 
\textbf{Surprisingly, we found the order of performance between PredKD, PredKD + EWC, and PredKD + L2 have been reversed!} While the forgetting metrics are not significantly affected, we see that $A_{1:N}$ has remained the same for PredKD but largely increased for EWC and L2. These results are reasonable after considering the following: with pre-training, less plasticity is needed because the model (features) are already useful for new tasks; thus, methods which achieved low forgetting at a cost of low performance on new tasks in the no pre-training scenario now have the ``cost" removed. For further validation of this perspective, we notice that the CKA similarity scores presented in Figure~\ref{fig:resnet18-pt_tentask_cka} have not changed much from Figure~\ref{fig:resnet18_tentask_cka}, yet the differential in performance on new tasks, presented in Figure~\ref{fig:resnet18_tentask_plasctiicy}, is strikingly  large. That is, pre-training does not seem to affect ``forgetting" for these methods but rather enhances the ability to learn new tasks without forgetting. \emph{In summary, the main takeaways from this section are that:} \textbf{1) L2 with PredKD outperforms EWC with PredKD in the presence of pre-training, and both of these approaches far outperform PredKD without parameter regularization} and \textbf{2) For parameter regularization methods, pre-training greatly improves the ability to learn new tasks but has little effect on forgetting.}

\subsection{Context with Current Literature - ResNet}
\begin{table}[t]
\caption{\textbf{Results (\%) for continual learning on CIFAR-100 on 10 tasks for different types of rehearsal and pre-training.} $A_{1:N}$ gives the final task accuracy.}
\centering

\begin{tabular}{c|c|c|c} 
\hline
\rule{0pt}{10pt} Method & Rehearsal & Pre-train & $A_{1:N}$ ($\uparrow$)  \\
\hline
Upper-Bound  & None & None & $ 56.2 $  \\ 
Naive  & None & None & $ 8.8 $ \\
\hline
\hline
PredKD  & None & None & $ 24.6 $ \\
PredKD + FeatKD  & None & None & $ 12.4 $ \\
PredKD + EWC  & None & None & $ 23.3 $ \\
PredKD + L2  & None & None & $ 21.5 $ \\

\hline
PredKD  & None & ImNet & $ 24.9 $ \\
PredKD + FeatKD  & None & ImNet & $ 21.7 $ \\
PredKD + EWC  & None & ImNet & $ 32.5 $ \\
PredKD + L2  & None & ImNet & $ \bm{34.4} $\\

\hline

DGR~\cite{he2016deep} & Gen. & None & $8.1$ \\
DeepInversion~\cite{yin2020dreaming} & Synth. & None & $10.9$ \\
ABD~\cite{smith2021abd} & Synth. & None & $33.7$ \\

\hline

Rehearsal & 2k IMG & None & $24.0$ \\
LwF~\cite{li2016learning} & 2k IMG & None & $27.4$ \\

\hline
\end{tabular}

\label{tab:context}
\vspace{-.1cm}
\end{table}

We compare our results with SOTA methods on the exemplar-free continual learning setting on CIFAR-100 using the saem 18-layer ResNet backbone. The difference between our setting of \emph{rehearsal-free} continual learning and this setting of \emph{exemplar-free} continual learning is that, while neither store images for rehearsal, the latter setting often synthesizes these images. As discussed in Section~\ref{sec:rl}, creating synthetic images with model inversion is a computationally expensive procedure and may still violate data-privacy concerns. We make our comparison in Table~\ref{tab:context}, showing final accuracy for the methods discusses in this paper. We find that \textbf{1) Pre-training can outperform SOTA rehearsal methods from synthetic data}, and \textbf{2) Pre-training can even outperform simple rehearsals methods that store a 2000 image coreset of data}.

\subsection{Context with Current Literature - ViT}

Next, motivated by our findings in the previous sections, we ask: \emph{can parameter regularization outperform prompting for continual learning methods?}~\cite{smith2022coda,wang2022dualprompt,wang2022learning}. Specifically, we conjecture that, given a fair implementation and comparison, which targets modifying only the same spot of the ViT model as prompting methods, parameter regularization might outperform prompting for these benchmarks.

We benchmark using ImageNet-R~\cite{hendrycks2021many,wang2022dualprompt} which is composed of 200 object classes with a wide collection of image styles, including cartoon, graffiti, and hard examples from the original ImageNet dataset~\cite{russakovsky2015imagenet}. This benchmark is attractive because the distribution of training data has significant distance to the pre-training data (ImageNet), thus providing a fair and challenging problem setting. We use the exact same experiment setting as the recent CODA-Prompt~\cite{smith2022coda} paper. We implement our method and all baselines in PyTorch\cite{paszke2019pytorch} using the ViT-B/16 backbone~\cite{dosovitskiy2020image} pre-trained on ImageNet-1K~\cite{russakovsky2015imagenet}. We compare to the following methods (the same rehearsal-free comparisons of CODA-Prompt): Learning without Forgetting (LwF)~\cite{li2016learning}, Learning to Prompt (L2P)~\cite{wang2022learning}, a modified version of L2P (L2P++)~\cite{smith2022coda}, and DualPrompt~\cite{wang2022dualprompt}. Additionally, we report the upper bound (UB) performance and performance for a neural network trained only on classification loss using the new task training data (we refer to this as FT). 

\looseness=-1
We freeze most of the backbone and only fine-tune the QKV projection matrices of self-attention blocks throughout the ViT model. The intuition is that we are modifying the same modules as the prompting methods, but using classic continual learning methods that fine-tune with regularization rather than add prompts. We use loss weights of $\{1e3,1\}$ for EWC and L2, respectively. Importantly, we use the same classification head as L2P, DualPrompt, and CODA-Prompt, and additionally compare to a FT variant, FT++, which uses the same classifier as the prompting methods and suffers from less forgetting. For additional details, we refer the reader to the CODA-Prompt~\cite{smith2022coda} paper. 

In Table~\ref{tab:imnet-r_main}, we benchmark against the popular and recent rehearsal-free continual learning methods. \emph{We found that L2 achieves a high state-of-the-art in this setting.} Compared to the prompting methods L2P, DualPrompt, and the recent CODA-Prompt, L2 has clear and significant improvements, whereas EWC has poor performance. Our intuition is that L2 is much stronger given it begins regularization in task 1 (rather than task 2, such as EWC), and regularizes not only for past tasks but also future tasks by encouraging the model parameters to stay close to rich initial pre-training state. \emph{In summary, the main takeaway from this experiment is that} \textbf{fine-tuning with L2 parameter regularization in the QKV projection matrices of ViT self-attention blocks outperforms prompting for continual learning methods.}
\begin{table}[t!]
\caption{\textbf{Results (\%) on ImageNet-R} for 10 tasks (20 classes per task, 3 trials). $A_{1:N}$ gives the final task accuracy and $F_N^G$ gives the average \emph{global} forgetting. We report mean \% stdev over 5 trials.
}

\label{tab:imnet-r_main}
\centering

\begin{tabular}{c|c|c} 
\hline
 \rule{0pt}{10pt} Method & $A_{1:N}$ ($\uparrow$) & $F_N^G$ ($\downarrow$) \\
\hline
UB
& $77.13$ & -   \\
\hline
FT        
& $10.12 \pm 0.51$ & $25.69 \pm 0.23$  \\
FT++       
& $48.93 \pm 1.15$ & $9.81 \pm 0.31$  \\ 
LwF.MC     
& $66.73 \pm 1.25$ & $3.52 \pm 0.39$  \\ 
L2P      
& $69.29 \pm 0.73$ & $2.03 \pm 0.19$  \\ 
L2P++ 
& $71.66 \pm 0.64$ & $1.78 \pm 0.16$  \\ 
DualPrompt
& $71.32 \pm 0.62$ & $1.71 \pm 0.24$   \\ 
CODA-P (small)
& $73.93 \pm 0.49$ & $1.60 \pm 0.20$  \\ 
CODA-P 
& $75.45 \pm 0.56$ & $1.64 \pm 0.10$  \\
\hdashline
EWC
& $64.66 \pm 2.04$ & $\bm{1.55 \pm 0.25}$   \\ 
\textbf{L2}
& $\bm{76.06 \pm 0.65}$ & $1.68 \pm 0.16$   \\ 

\hline
\end{tabular}
\vspace{-.1cm}
\end{table}
\section{Conclusions}
In this work, we take a closer look at several popular continual learning strategies in the setting of \emph{rehearsal-free continual learning}. This setting reflects machine-learning applications which cannot store or generate past-seen training data due to privacy concerns or memory constraints. We first show that parameter regularization techniques such as L2 and EWC can succeed in the rehearsal-free continual learning setting if softmax is removed from the classification head. Then, we compare parameter regularization, feature distillation, and prediction distillation on a 10-task continual learning benchmark. We find that with a randomly initialized model, parameter regularization methods achieves low forgetting but at the cost of low accuracy. When we initialize the model with pre-trained weights, we find that parameter regularization injects both low forgetting \emph{and} high accuracy. Surprisingly, we found that \emph{L2 regularization outperforms EWC in the pre-trained model scenario}. To validate these findings, we demonstrate that L2 parameter regularization implemented in a ViT transformer outperforms recently popular prompting for continual learning methods. In conclusion, our study has provided valuable insights into the efficacy of different types of regularization for continual learning and highlighted the potential of regularization in rehearsal-free settings. 

\section*{Acknowledgements}

This material is based upon work supported by the National Science Foundation under Grant No. 2239292.

{\small
\bibliographystyle{ieee_fullname}
\bibliography{references}
}

\end{document}